\documentclass[11pt,a4paper]{article}
\usepackage[hyperref]{acl2020}
\usepackage{times}
\usepackage{latexsym}
\usepackage{amsmath}
\usepackage[utf8]{inputenc}
\usepackage{graphicx}

\newcommand\numberthis{\addtocounter{equation}{1}\tag{\theequation}}

\usepackage{microtype}

\aclfinalcopy 

\title{Normalizing Text using Language Modelling based on Phonetics and String Similarity}

\author{Fenil Doshi\textsuperscript{1}\thanks{Equal Contributors}, 
Jimit Gandhi\textsuperscript{2}\footnotemark[1], 
Deep Gosalia\textsuperscript{3}\footnotemark[1] \and
Sudhir Bagul\textsuperscript{4} \\
  \textsuperscript{1,2,3,4}Dwarkadas J. Sanghvi College of Engineering, Mumbai \\
  \texttt{\textsuperscript{1}fenildoshi25@gmail.com  \textsuperscript{2}jimitgandhi@outlook.com} \\
   \texttt{\textsuperscript{3}deepgosalia1@gmail.com  \textsuperscript{4}sudhir.bagul@djsce.ac.in} \\  }

\date{}

\begin{document}
\maketitle
\begin{abstract}
Social media networks and chatting platforms often use an informal version of natural text. Adversarial spelling attacks also tend to alter the input text by modifying the characters in the text. Normalizing these texts is an essential step for various applications like language translation and text to speech synthesis where the models are trained over clean regular English language. We propose a new robust model to perform text normalization. 

Our system uses the BERT language model to predict the masked words that correspond to the unnormalized words. We propose two unique masking strategies that try to replace the unnormalized words in the text with their root form using a unique score based on phonetic and string similarity metrics.We use human-centric evaluations where volunteers were asked to rank the normalized text. Our strategies yield an accuracy of 86.7\% and 83.2\% which indicates the effectiveness of our system in dealing with text normalization.

\end{abstract}

\section{Introduction}
Today, the internet and SMS has become a universal platform for people to communicate. People alter the natural text in various ways to communicate using SMS language. These alterations make it it's own separate language in a way, except for one key aspect: the lack of consistency. Each person may choose to eliminate a different vowel or in general, use a separate form to represent one specific word.

Apart from this, one of the prominent black-box adversarial attacks on text is character-level adversarial attacks where the letters in a word are randomly added, deleted or swapped. Such types of perturbations significantly reduce the accuracy of Deep learning models as shown in the paper: \citep{Danish18} where a character-level modification drops the accuracy on the sentiment-analysis model fine-tuned on BERT from 90\% to 45.8\%. These alterations are completely random without any regularity.    

The lack of consistency in such methods indicates that a rule-based approach will not yield good results. Every sentence has its own unique representation that cannot be quantified by a set of rules. 

Text Normalization becomes an essential part of any NLP pipeline. Most of the data mined from the web is not in a consistent English format that the language processing models are trained on. Text normalization forms the process of converting such informal text into a suitable standard format. In this paper, we propose a unique model to perform this task. Our model considers the context of the word in use along with other parameters. Most of the other papers, to the best of our knowledge, follow traditional approaches without considering the context in which the word is being used.
We also use a unique similarity metric that incorporates a combination of string similarity, phonetic similarity as well as the probability of occurrence of the word to predict the root form of the unnormalized word. 

We propose two separate approaches to tackle the problem. The first one requires a less computational effort but fails to handle cases where the unnormalized word is also a part of the dictionary. The second approach handles this case by making a prediction for every word in the sentence but takes more time for execution.

Most papers tackling this problem use traditional machine translation approaches and hence their evaluation metrics are not pertinent to our approach. We use a human-centric approach for evaluation where volunteers were asked to rank the normalized text and provide a comprehensive analysis of both our methods.

\section{Literature Review}
With the current trend of Social Media and online chatting applications, it becomes imperative that various Text Normalisation methods have been researched and applied in order to determine an efficient system for normalization. The previous work in the field mainly comprises methods like Machine Translation, Rule-Based approach, models based on phonetics and string similarity metrics or a combination of these. 

One paper: \citep{Aw1} viewed the problem of SMS normalization as a translation problem and used phrase-based statistical machine-translation based on the work of another paper: \citep{Koehn6}. However, their model failed to handle missing punctuations and did not use information from pronunciations for OOV (Out-Of-Vocabulary) words.

Another approach: \citep{Raghunathan3} used a dictionary substitution method and an off-the-shelf machine translation to tackle the given problem. Their BLEU score was better than the paper: \citep{Aw1}, because of the various deletions made by the paper: \citep{Aw1}

The following work: \citep{Yang-Liu5} used both word-level and sentence-level optimization schemes using two different methods: creating and using a reverse lookup table and Word Level Reranking. Due to the evaluation of only a small set of word candidates, their reranking provided better feasibility in the normalization step. 

Two-phase expansion method of abbreviations was used by: \citep{pennell11}, using Machine translation at character-level and an in-domain language model. While their work procured distinguished results when the character-level MT was integrated with contextual information at the word level, they had assumed their target words to be abbreviations only. To handle the shortcomings of MT-based models, the authors of the paper: \citep{Gadde7} proposed a controlled method to add artificial noise to regular English corpora to effectively test the MT-based approach.

A hybrid strategy towards SMS text normalization was proposed by: \citep{Meenakshi2}, using a Rule-Based, Direct Mapping and Machine Translation approach. Her system was targeted at normalizing SMS text before the process of Machine Translation, thus eliminating the tremendous effort required to adapt the language model of the existing system to handle that particular SMS text style.

Another paper: \citep{khanuja4} proposed an implementation that used morphophonemic and lexical similarity to generate the set of correcting candidates that could contain a replacement for the OOV word. Similarly, the paper: \citep{Pinto13} performed the task on both, English and Spanish languages by employing the Soundex algorithm and a modification of it. They also used Jaccard distance as a similarity measure for evaluating the different adaptations of proposed Soundex algorithms.

Additionally, the paper: \citep{Han12} used both contextual and string similarity information. With a focus on context-insensitive lexical variants, their type-based normalization was able to achieve good accuracy with reasonable precision.

\section{Characteristics of SMS Language}
The main issue that arises with the given problem statement is the varying nature of SMS Language. The lack of syntactical rules means that the usage of SMS language can vary from person to person and the same word can be represented in several forms. Consider the word 'where' which might be used and 'whr' and 'wher' or may have some other representation altogether. But, even with the inconsistencies, the overall purpose of using this form of text is common and can be categorized in the form of the given variants:

\noindent
(i) Repetition of letters of a word to stress the importance (cool $\rightarrow$ coooooool)

\noindent
(ii) Eliminating non-essential letters in a way that the word phonetically still remains the same and the semantics can be extracted from shortened word (friend $\rightarrow$ frnd) 

\noindent
(iii) Replacing letters or a group of letters by other similar-sounding syllables(tomorrow $\rightarrow$ 2morrow) 

\noindent
(iv) Using common text abbreviations for frequently used phrases (Laughing out loud $\rightarrow$ lol) 

\noindent
(v) Not respecting grammatical syntax rules for capitalization as long as sentence conveys the essential meaning (Mr. Barack Obama $\rightarrow$ mR barack obAMa) 

\noindent
(vi) Lack of punctuation (can't $\rightarrow$ cant) 

\noindent
(vii) Excessive punctuation for emphasis (Why? $\rightarrow$ why????)

\noindent
(viii) Use of contractions (you have $\rightarrow$ you've)

\section{Problem Formulation and Assumptions}
\label{pf}
Before tackling the process of normalization, we address several problems. Firstly, several languages may be used while communicating using text, but our model focuses only on the English language. Next, the various informalities mentioned in the previous section need to be normalized. Even after doing these, several anomalies need to be addressed. One possible case is when an unnormalized word is not necessarily an OOV word but is itself an English word from the dictionary. For instance,

\noindent
Eg. 1) "I am \underline{with} her" $\rightarrow$ "i m \underline{wit} her."

Here, the word "wit" itself exists in the dictionary. However, here, it is used as the unnormalized form of the root word "with".

\noindent
Eg. 2) "i m \underline{bout} 2 jmp in d river." 

\noindent
Eg. 3) "\underline{Cud} u pass me d glass of water?"

Another possible case depends on the context in which the word is used. The same informal word can be used to represent two different words in two different sentences. For example,

\noindent
Eg. 4) "The train leaves in 5 \underline{min}" $\rightarrow$ "The train leaves in 5 \underline{minutes}."

\noindent
Eg. 5) "\underline{Min} 1 paper is required" $\rightarrow$ "\underline{Minimum} 1 paper is required."

Similarly, An informal word may be an abbreviation or a Named-Entity: 

\noindent
Eg. 6) "It's hot \underline{bc} it's humid" $\rightarrow$ "It's hot \underline{because} it's humid."

\noindent
Eg. 7) "I m living in \underline{bc}" $\rightarrow$ "I am living in \underline{British Columbia}."

Our proposed approach assumes that the grammar of the input sentence is accurate and makes no effort to alter the grammatical correctness. Consequently, if some grammatically incorrect sentence is encountered then the normalization may yield less accurate results. We also do not treat upper case characters separately. In the output sequence, all words, including the first word of the sentence have lower cases only. Additionally, smileys and emojis ( :-), :D, :-P ) are not separately handled by our model and our treated as separate punctuation and/or characters. Lastly, our model assumes that each word is separated by a space. If two words are concatenated together then they will be treated as one whole word. 

\noindent
Eg. 8) "I will \underline{missyou}."

Our model also performs much better when the informality ratio of the text is low to medium. We define informality ratio as:
\begin{align*}
    informality\_ratio &= n/N \numberthis
\end{align*}
where n is the number of unnormalized informal words and N is the total number of words in the text.

\section{Our Model}
\begin{figure}[t]
    \centering
    \includegraphics[width=7.5cm]{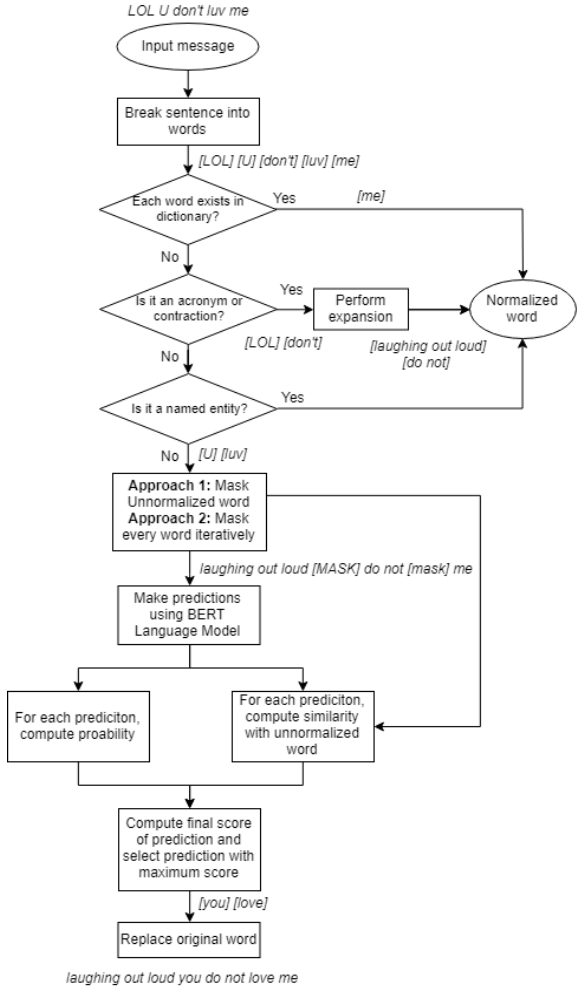}
    \caption{Flowchart of Proposed Model}
    \label{fig-flowchart}
\end{figure}

We propose a hybrid unsupervised approach for dealing with unnormalized text. We present a new pipeline architecture that models the context of the word before converting it to its normalized form. We propose two different strategies for this purpose. The flow of both these approaches is shown in Fig. \ref{fig-flowchart}. The first approach follows the masking of only OOV words (except named entities, acronyms, contractions) while the other approach follows masking of every word. The steps of the proposed system are as follows:

\subsection{Approach 1: OOV Masking}
\label{app1}
\subsubsection{Tokenization}
The input corpus of informal text is first broken down into individual sentences. Each individual sentence is processed independently and is further tokenized to produce a list of words. Each word then undergoes a series of steps in order to determine its form.

\subsubsection{Detecting Unnormalized tokens}
This step is processed over individual words (except Named Entity Recognition). We classify each word into five categories and handle each class individually.

(a) \textbf{Normalized words:} This forms the major portion of the text and these words should be left untouched.  (e.g.- "Hello", "cricket", "language", etc.)

(b) \textbf{Acronyms:} A short-form of prominent text phrases frequently used in text. (e.g.- "LOL", "GM", "FML", etc.)

(c) \textbf{Contractions:} A word created by shortening and merging two different words (e.g.- "can't", "haven't", "I'll", etc.)

(d) \textbf{Named Entities:} These are generally the nouns (names, organizations, places, etc.) present in the text and shouldn't be modified. (e.g.- "Jack", "Baltimore", etc.)

(e) \textbf{Unnormalized tokens:} These form the list of all tokens where the root form of the token (normalized word) is altered by the user based on some modifications (random additions, deletions, substitutions, transpositions, etc.) such that the new word effectively conveys the same meaning. (e.g.- "coooool", "frndshp", "al2gether", etc.)

These categories are not mutually exclusive and there can be significant overlap over each other. (e.g.- "wit", "Will", "US", "couldnt", etc.). However, our first approach assumes them to be disjoint sets. We identify each word by going through the following series of steps. Since each set is considered to be mutually exclusive, the sequence of steps is immaterial and can be executed in any order.

\textbf{Step I - Dictionary LookUp}

We create an exhaustive vocabulary where words from the website: \citep{Corncob10} are unioned with a list of words provided by the NLTK library to form a vocabulary of \textbf{59493} words. Each word's presence is then checked in the vocabulary. If the word exists, then we leave the word as it is and move to the next word in the list of tokens (since we assume it to be in it's the correct normalized form).
All the other words are OOV (Out-of-Vocabulary words) and fall into one of the other four categories.

\textbf{Step II - Acronym Matching}

Step II and Step III are based on a rule-based matching approach. The word is searched through a robust table of prominent acronyms table and if such an acronym exists, then it is replaced by its expansion. All popular internet acronyms are included in the table for effective expansions. 
A sample of the table is shown in table \ref{acronym-expansion}.

\begin{table}
\centering
\begin{tabular}{ll}
\hline \textbf{Acronym} & \textbf{Expansion} \\ \hline
GM & Good Morning \\
ROFL & Rolling On the Floor Laughing \\
AWOL & Absent Without Leave \\
CYA & See you \\

\hline
\end{tabular}
\caption{\label{acronym-expansion} Acronym Expansion }
\end{table}

The table contains 321 such expansions.

\textbf{Step III - Contraction Matching}

The above table is modified and additionally, 120 other entries are also added that provide expansions for contractions in the English language making the final length of the table as 441. A sample of contractions and expansions is shown in Table \ref{contractions}.
All the contractions/acronyms are searched through this final table.

\begin{table}
\centering
\begin{tabular}{ll}
\hline \textbf{Contraction} & \textbf{Expansion} \\ \hline
we're & We are \\
she'll & She will \\
hasn't & Has not \\
you've & You have \\
\hline
\end{tabular}
\caption{\label{contractions} Contraction Expansion }
\end{table}

\textbf{Step IV - Named Entity Recognition}

Not all OOV (out-of-vocabulary) words need to be modified. Some of them are proper nouns that are already in their root form and need to be kept unchanged. For this purpose, we recognize such named entities in the text by using NER (Named Entity Recognition). Most of the existing approaches use Gazetteers for this task. Such approaches require the text to be in a normalized form where the accurate grammatical syntax is followed. (All the proper nouns should have their first letter capitalized. e.g- Will, Adam, etc.). Such rules are not followed in the informal language where any number of letters in a word can be randomly capitalized. (e.g- WiLL, ADAM, etc.)

Hence, for NER on informal text, we adopt a hybrid model using Gazetteers and a fine-grained residual bidirectional LSTM network trained on ELMo embeddings for the NER task, as in: \citep{Matt14}. The performance of the fine-grained model depends upon the informality ratio of the text. Higher the ratio, the less accurate is the prediction of the model because the model is trained on formalized English corpus whereas the downstream task is predicting it on informal Text. Therefore, we use the model along with a gazetteer to curate named entities in the text.

This step takes into consideration the entire text and not just a single word, unlike previous steps. All the words that are recognized as names, organizations, places and other named entities from the model are left untouched as they are in their root form and need not be updated. 

\textbf{Step V - Unnormalized Words}

We assume that the words which are not categorized in any of the above four classes are unnormalized words. These are the set of words that need to be converted into their root form.

\subsubsection{Converting unnormalized words into their root forms}
This task involves converting unnormalized words into their normalized form while considering the context. Here, we use Masked Language Modelling along with similarity metrics to find the root form of the word.
It involves the following steps:

\textbf{(i) Finding all the possible replacements of the unnormalized words based on its context}

The previous task returns a list of unnormalized words. Each of these given words is then masked in the sentence (The token is replaced by '[MASK]' token). We use a pre-trained BERT language model \citep{Jacob15} to predict these masked tokens. BERT is a state-of-the-art transformer model that is trained on language modeling tasks using bidirectional transformers so that it can model context from both directions. It is particularly useful for our task because the context of the masked word needs to be considered from both directions, forward as well as backward.  

We use the 'BERT-base-uncased model' as defined by the paper since the casings of the letters hold no value for informal text. The sentence is then converted into tokens using the WordPiece Tokenization, and then the tokens are fed to the trained BERT language model. The model returns a list of probable predictions. The accuracy of the predictions is also dependent upon the informality ratio where it performs much better in a low to moderate setting. This is also because the BERT model was trained on formal English text and performs poorly when the source text is modified. 

The model outputs a list of possible predictions along with their scores based on their probability. These scores are unevenly distributed. We sort the list of predictions in descending order, based on these scores. We restrict this list to the top 5000 words since we observe that all the possible root words are within the initial 5000 words, thus leading to faster normalization. 

\textbf{(ii) Computing a unique score to find the corresponding root word in the list}

We compute a unique score for each possible word in the above list of 5000 words with respect to the unnormalized word. This score consists of 3 components - 

\textbf{(a) Probability based on context:} The output score of the model after the softmax layer is non-linearly distributed and is biased towards the initial few predictions. This is undesirable because the root form of the word need not be in the top few predictions but also has a considerable chance of occurring later. Thus, we consider a simple probability metric where we assume a linear distribution over all the possible predictions in the list. It is defined as:
\begin{align*}
    P(X) &= 1 - (index(X)/5000) \numberthis \label{eqn-prob}
\end{align*}
where X is one of the possible root words in the prediction list and index(X) is the index of the word in the prediction list of length 5000 (sorted in descending order)

All the other words in the list (after 5000) are considered to have a probability of 0 and are not considered further.

\textbf{(b) String Similarity:} Each word in the shortened list is then scored based on the string distance between the word and its unnormalized form. The string similarity is computed using Levenshtein, Jaro-Winkler and Cosine similarity between the two words. Levenshtein helps in handling insertions and deletions whereas Jaro-Winkler additionally deals with transpositions among the words. Cosine similarity is computed between the two sets of unigrams and bigrams of predicted and unnormalized words. All the values are normalized between 0 and 1.
\begin{align*}
    SSim(X, Y) =& 0.6*NL(X,Y) + \\
    & 0.2*JW(X,Y) + \\
    & 0.15*C(X1,Y1) + \\
    & 0.05*C(X2,Y2) \numberthis
\end{align*}
where X is one of the possible root words in the prediction list, Y is the unnormalized word, SSim(X,Y) is the String Similarity between words X and Y, NL(X,Y) is the Normalized Levenshtein Similarity between string X and Y, JW(X,Y) is the Normalized JaroWinkler Similarity between string X and Y, Xn is the Set of n grams of word X and C(Xi,Yi) is the Cosine Similarity between Sets Xi and Yi.

The coefficients in the equation are decided by the relative importance of individual measures using experimentation.

\textbf{(c) Phonetic Similarity:} In the Text Normalization process, since the root word is definitely phonetically similar to its unnormalized form, the phonetic similarity is more valuable than string similarity. First, we substitute all the non-alphabetical symbols in the unnormalized word with their phonetically similar-sounding syllables based on their usage in SMS language (e.g. '4' is replaced with "for", '@' is replaced with "at", etc.) 
We then compute the phonetic similarity between the two words using Soundex, Metaphone and Fuzzy Soundex algorithms. The string is converted into its phonetic code, followed by calculating normalized Levenshtein similarity between the two codes.
\begin{align*}
     PSim(X,Y) =& 0.6*NL(M(X), M(Y)) + \\
     & 0.2*NL(S(X), S(Y)) + \\
     & 0.2*NL(FS(X), FS(Y)) \numberthis
\end{align*}
where M(X) is the Phonetic code of word X after applying Metaphone algorithm, S(X)is the Phonetic code of word X after applying Soundex algorithm, FS(X) is the Phonetic code of word X after applying Fuzzy Soundex algorithm, NL(X,Y) is the Normalized Levenshtein Similarity between codes X and Y and PSim(X,Y) is the Phonetic Similarity between words X and Y.

The Similarity Score is then computed using Phonetic and String Similarity with the following weights:
\begin{align*}
    SimScore(X,Y) =& 0.65*PSim(X,Y) +  \\
    & 0.35*SSim(X,Y) \numberthis
\end{align*}
Most of these words are more phonetically similar to each other than their string counterparts. Hence, Phonetic Similarity is given more weightage over String Similarity. 

"The Cambridge Effect" demonstrates that jumbled letters have little difference on reading as long as the first and last letter of the word remains the same. This is frequently observed in the case of SMS text. Thus, we award the Similarity score when the first letter and last letter of the unnormalized word and the prediction word matches. We intensify the score in such cases, whereas, if both the letters don't match, we dilute the Similarity Score.
\begin{align*}
    SimScore(X,Y) = SimScore(X,Y)^2 \\
    (if X[0]=Y[0]\ and\ X[-1]=Y[-1]) \numberthis
\end{align*}
\begin{align*}
    SimScore(X,Y) = SimScore(X,Y)^{0.5} \\
    (if X[0] \neq Y[0]\ and\ X[-1] \neq Y[-1]) \numberthis 
\end{align*}
where X[0] is the First letter of predicted word X, Y[0] is the First letter of unnormalized word Y, X[-1] is the Last letter of predicted word X and Y[-1] = Last letter of unnormalized word Y.

This Similarity Score is then considered along with the Probability to compute the final score associated with each predicted word in the prediction list. We simply multiply them to get the final score associated with each word in the list.

\begin{align*}
    FinalScore(X) &= P(X)*SimScore(X,Y) \numberthis
\end{align*}

\textbf{(iii) Final Step of replacing each word based on the Score:} All the predicted words have an associated final score and the word having maximum final score is selected as a replacement and is predicted to be the root word of its unnormalized form.

However, if the best FinalScore is less than 0.25, then we assume that none of the best predictions are the root form of the word and we leave the unnormalized word as it is.

\subsection{Approach 2: Word-by-Word Masking}
One of the major limitations of the previous approach was the assumption that all unnormalized words are OOV words (other than Named entities and Acronyms) that need to be normalized. However, such is not the case as discussed in Section \ref{pf}. Some words need to be normalized even though their unnormalized form is present in the dictionary. (e.g- "wit" for "with", "cud" for "could", etc.). Approach 1 couldn't handle such cases and hence, we propose Word-by-Word masking as an alternative approach.

Out of the 5 categories discussed in Section \ref{app1}, Approach 1 just masked words in the fifth category ("Unnormalized words") whereas this strategy masks all the words in the fifth as well as first category (both "Normalized" and "Unnormalized" words).

The pipeline architecture described in the previous approach is followed here also, with the same steps and formulae except for the Masking step. Here, each word that is not an acronym, contraction or a named entity is masked and is then predicted by the Masking model. We assume that if the word is already in the root form then, the predictions returned from the masking model will have the word in it's top few predictions. The similarity score of the word with itself is always 1. Hence, the model should encounter no problems if the word is already in its normalized form.

However, if the token is present in the dictionary and still is unnormalized, then the prediction list returned by the model will have its root form in the earlier predictions and the word itself in the rear part of the list. Even though, the similarity with an unnormalized word will be 1, the probability (P(X)) term in Eq.\eqref{eqn-prob} will be quite low and such words will be penalized. The root form, as it is in the front of the list will have a high probability as well as high similarity. Even though similarity won't be exactly equal to 1, it will be considerably higher and when combined with the probability, will yield a higher FinalScore then the unnormalized form of the word. 

One of the caveats of this method is the time taken to compute the final output of the sentence. Since almost all the words are masked, the total number of times the model makes its predictions equals the length of entire sentences. On the other hand, total predictions in Approach 1 approximately equals the length of sentence multiplied by the informality ratio. Hence, the higher the number of times the model makes a prediction, the more is the compute time required to find the normalized output.

We present our findings from both these approaches and give a comparative analysis of which method is better.

\section{Experimental Results}

\begin{table*}[t]
\centering
\begin{tabular}{lcccccc}
\hline \textbf{Approach}  & \textbf{1} & \textbf{2} & \textbf{3} & \textbf{4} & \textbf{5} & \textbf{Accuracy} \\ \hline
Approach 1(OOV Masking) & 43 & 89 & 292 & 722 & 1481 & \textbf{86.71\%} \\
Approach 2(Word-by-Word Masking) & 41 & 133 & 464 & 713 & 1276 & \textbf{83.22\%} \\
\hline
\end{tabular}
\caption{\label{tab-app1} Number of Ratings and Accuracy for Each Approach }
\end{table*}

To measure the accuracy of our model, we follow a human-centric evaluation approach. We found this approach to be the most accurate way to analyze if the meaning of the word is retained after normalization because of a lack of a well-annotated dataset.

For analysis, we use a corpus of SMS texts collected by: \citep{Almeida16}. This dataset consists of 5,574 SMS messages. We eliminate the messages classified as "SPAM" and run the rest of them (about 2627 messages- having class as "HAM") through our model using the first approach, where predictions are made only on the unnormalized words. The total runtime was 260 minutes, averaging at around 5.9 seconds per message using a Tesla K80 GPU (12GB RAM). We also ran the same set of messages through the model using the second approach, where every word was treated as an unnormalized word in the sentence. As predicted, it took considerably more runtime: 450 minutes, averaging at about 10.2 seconds per message.

A total of 2627 messages were evaluated by the volunteers. Each text message contains an average of about 18 words. Out of these, 185 messages in Approach 1 and 133 messages in approach 2 underwent no change after passing through the model since these messages were already in their normalized form, i.e. Having the informality ratio as 0. 

To perform an evaluation on this data, we gathered over 30 volunteers and provided them with a set of input and output texts. They were instructed to rate the accuracy of normalization of the output messages on a scale of 1-5 where 1 is very inaccurate and 5 is most accurate. To prevent any bias, volunteers were either given the outputs of the first approach or the second one, not both.

We average out the ratings provided to each tuple and compute the final accuracy of the model with the following formula:

\begin{align*}
    Accuracy=(20/N)*\sum_{i=1}^{N}r_i \numberthis
\end{align*}
where r\textsubscript{i} is the Average rating of tuple i for 1$\leq$r\textsubscript{i}$\leq$5 and N is the Total number of tuples.

The results obtained after the evaluation are shown in table \ref{tab-app1}
\begin{figure}[h]
    \centering
    \includegraphics[width=7.5cm]{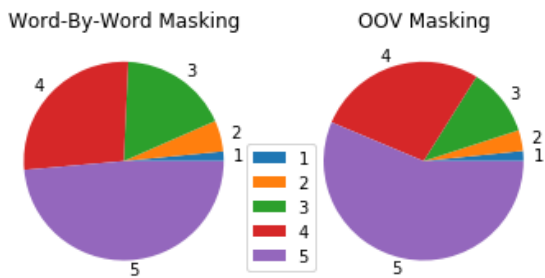}
    \caption{Pie chart of ratings of both approaches}
    \label{fig:pie1}
\end{figure}
We observe that both approaches provide competitive results. Approach 1 yields an accuracy of 86.71\%,  slightly better than the accuracy of Approach 2 of 83.22\%, even though it requires comparatively less prediction time. 
The results were also independent of the length of the sentences. The distribution of ratings was similar for bigger as well as smaller sentences, indicating that the model could effectively model the context even in longer sequences. 
\begin{figure}[h]
    \centering
    \includegraphics[width=7.5cm]{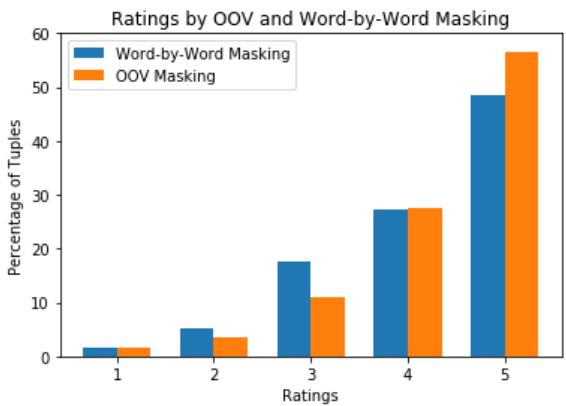}
    \caption{Comparison of Ratings between Approach 1 and Approach 2}
    \label{fig:bar}
\end{figure}
We assume that comparatively lower ratings of Approach 2 were due to it's masking of every word. While this helps in dealing with unnormalized words that are a part of the vocabulary, such words form a very small part of the text. While dealing with them, this approach sometimes falsely modifies the correct root forms of the word and changes them to some other probable word that has high similarity to the original word. This reduces the effectiveness of conversion. While such cases are very few, they are not present in OOV word masking and hence, it provides slightly better results then Word-by-Word masking at the cost of ignoring all unnormalized dictionary words.

\section{Future Work}
We primarily intend to improve the quality of the normalization process in future work. Since, the current models used, like BERT and AllenNLP, are pre-trained on the normalized English text, and not informal text, they provide average results during the prediction. The issue of unavailability of annotated data is also a key factor due to which training of these models becomes difficult. In the future, we aspire to train these models, for Named Entity Recognition (NER) using a surfeit of annotated data derived from our informal text. The masking strategy either masks all words or only the OOV words. With the availability of annotated data in the future, we can also devise a joint probabilistic distribution over data which indicates which words to mask and thus, reducing the total number of passes through the masking model which consequently, reduces the prediction time.

With our currently designed model, when two different informal words are combined together and provided as an input, the model fails to correctly mask them. For eg. for the words 'Gonnamissu' our model masks it as 'gaining'. Hence, we would also handle the instances where more than one word is combined, resulting in the creation of a new OOV informal word.

\section{Conclusion}
We have proposed two different approaches in order to deal with unnormalized text. These systems find their use in converting SMS text to formal language, combating adversarial character-level spelling attacks, etc. We evaluate each procedure's performance on a corpus of English informal messages. We observe that both the approaches provide competitive results in dealing with such text and are effectively able to perform the process of normalization.

\bibliographystyle{acl_natbib}
\bibliography{acl2020}

\end{document}